\documentclass[conference]{IEEEtran}

\IEEEoverridecommandlockouts

\usepackage{cite}
\usepackage{amsmath,amssymb,amsfonts}
\usepackage{algorithmic}
\usepackage{textcomp}
\usepackage{xcolor}
\usepackage{booktabs}
\usepackage{threeparttable}
\usepackage{float}
\usepackage{mathtools}
\usepackage{lipsum}
\usepackage{amsmath,amssymb,amsfonts}
\usepackage{graphicx}
\usepackage{subcaption}
\usepackage{todonotes}
\usepackage{multirow}
\usepackage{soul}
\usepackage{tabularx}

\usepackage[absolute,showboxes]{textpos}
\TPMargin{5pt}
\newcommand{\copyrightstatement}{
\begin{textblock}{0.84}(0.08,0.93)    
\noindent
\footnotesize
\copyright 2020 IEEE. Personal use of this material is permitted. Permission from IEEE must be obtained for all other uses, in any current or future media, including reprinting/republishing this material for advertising or promotional purposes, creating new collective works, for resale or redistribution to servers or lists, or reuse of any copyrighted component of this work in other works.
\end{textblock}
}

\def\BibTeX{{\rm B\kern-.05em{\sc i\kern-.025em b}\kern-.08em
    T\kern-.1667em\lower.7ex\hbox{E}\kern-.125emX}}

\begin{document}

\title{Transformation Based Deep Anomaly Detection in Astronomical Images
\thanks{Esteban Reyes acknowledges financial support from the National Agency for Research and Development (ANID) /  Scholarship Program / MAGISTER NACIONAL/2019 - 22190947. The authors acknowledge financial support from ANID-Chile through grant FONDECYT 1171678. Also, the authors acknowledge support from the Chilean Ministry of Economy, Development, and Tourism's Millennium Science Initiative through grant IC12009, awarded to the Millennium Institute of Astrophysics, MAS. Additionally, the authors acknowledge financial support from the Department of Electrical Engineering at Universidad de Chile.}
}

\author{
\IEEEauthorblockN{Esteban Reyes}
\IEEEauthorblockA{
\textit{Department of Electrical Engineering}\\
\textit{Universidad de Chile}\\
Santiago, Chile \\
 esteban.reyes@ug.uchile.cl}
\and
\IEEEauthorblockN{Pablo A. Est\'evez}
\IEEEauthorblockA{
\textit{Department of Electrical Engineering}\\
\textit{Universidad de Chile}\\
Santiago, Chile \\
pestevez@cec.uchile.cl}
}

\maketitle

\begin{abstract}
In this work, we propose several enhancements to a geometric transformation based model for anomaly detection in images (GeoTranform). The model assumes that the anomaly class is unknown and that only inlier samples are available for training. We introduce new filter based transformations useful for detecting anomalies in astronomical images, that highlight artifact properties to make them more easily distinguishable from real objects. In addition, we propose a transformation selection strategy that allows us to find indistinguishable pairs of transformations. This results in an improvement of the area under the Receiver Operating Characteristic curve (AUROC) and accuracy performance, as well as in a dimensionality reduction. The models were tested on astronomical images from the High Cadence Transient Survey (HiTS) and Zwicky Transient Facility (ZTF) datasets. The best models obtained an average AUROC of 99.20\% for HiTS and 91.39\% for ZTF. The improvement over the original GeoTransform algorithm and baseline methods such as One-Class Support Vector Machine, and deep learning based methods is significant both statistically and in practice.

\end{abstract}

\copyrightstatement

\section{Introduction}
Astronomy has entered the era of big data as a result of the construction of large facilities such as the Zwicky Transient Facility (ZTF) \cite{dekany2020zwicky} and the Large Synoptic Survey Telescope (LSST) \cite{ivezic2019lsst}. ZTF is a telescope that is currently generating about 1 million astronomical alerts per night. On the other hand, the LSST, which will begin operations in northern Chile in 2022 \cite{huijse2014computational}, will collect information of over 10 million astronomical objects every night. Both telescopes generate alerts of astronomical objects that change in time or position, e.g. supernovae (SNe), explosions of dying stars. To process all these alerts, intermediary agents called \emph{astronomical alert brokers} are under development to fulfill the tasks of receiving, processing, classifying and reporting the identified objects, in order to facilitate their study by the astronomical community. The present work has been developed as a part of the ALeRCE (Automatic Learning for
the Rapid Classification of Events) broker which is currently processing ZTF data in preparation for the LSST era. Telescope alerts come in triads of images: the first observation of an object (\emph{template}), a posterior observation (\emph{science}) and the \emph{difference} image generated through a subtraction-like process between the template and science images. Many of the alerts produced by a telescope are not true objects but artifacts, caused by misalignment between template and science images that result in a bad subtraction in the difference image, background fluctuations, and defective CCD pixels, among others. In the literature, most works have focused on the task of automatically filtering out artifacts using a supervised approach,  i.e., they use a group of experts to manually label alerts as \emph{real} objects or \emph{artifacts} (also known as \emph{bogus})\cite{cabrera2016supernovae}, \cite{cabrera2017deep}, \cite{reyes2018enhanced},
\cite{duev2019real}. In our experience with the ZTF alert stream for over a year, new types of artifacts continue appearing due to the variable nature of these events. Up to now, a group of astronomers is manually labeling the bogus every day, which takes both time and a lot of effort. In this paper, we aim at detecting bogus automatically, but without using this class of events in the training set.

We assume that \emph{only} the inliers (real transient alerts) are known, which in astronomy can be obtained through the process of cross-matching \cite{zhao2009paralleled} with other catalogs, while the anomalies (bogus alerts) are completely unknown and are only available in the test sets. Using the inliers only to train anomaly detectors, is known in the machine learning literature as \emph{one-class anomaly detection}. In \cite{lin2018machine}, an approach to detect bogus events using real objects as inliers to train an Isolation Forest (IF) model \cite{liu2008isolation} was proposed, but they used handcrafted features instead of the images directly. 

We propose an enhancement of a geometric transformation based anomaly detection model (GeoTransform) \cite{golan2018deep}, a state-of-the-art algorithm. GeoTransform performs a series of geometric transformations over the images of inliers, and a classifier is trained to discriminate among these transformations. Anomalies are expected to be miss-classified. The two main contributions of our work are the following. First, we include Gaussian and Laplacian filter transformations that highlight features of astronomical artifacts. Second, we propose a novel selection strategy, which eliminates useless transformations, reducing the need to calculate redundant transformations and lowering the complexity and dimension of the classification space. We use a-priori knowledge that astronomical images are rotational invariant because the objects of interest are point sources and the angle from where they are observed does not greatly affect its appearance.  We compare these proposed enhancements to GeoTransform with several baseline algorithms for anomaly detection, achieving state-of-the-art results for astronomical artifact detection.

\section{Related Work}

\subsection{High Cadence Transient Survey}

\begin{figure}[t]
\centerline{\includegraphics[height=0.23\textwidth]{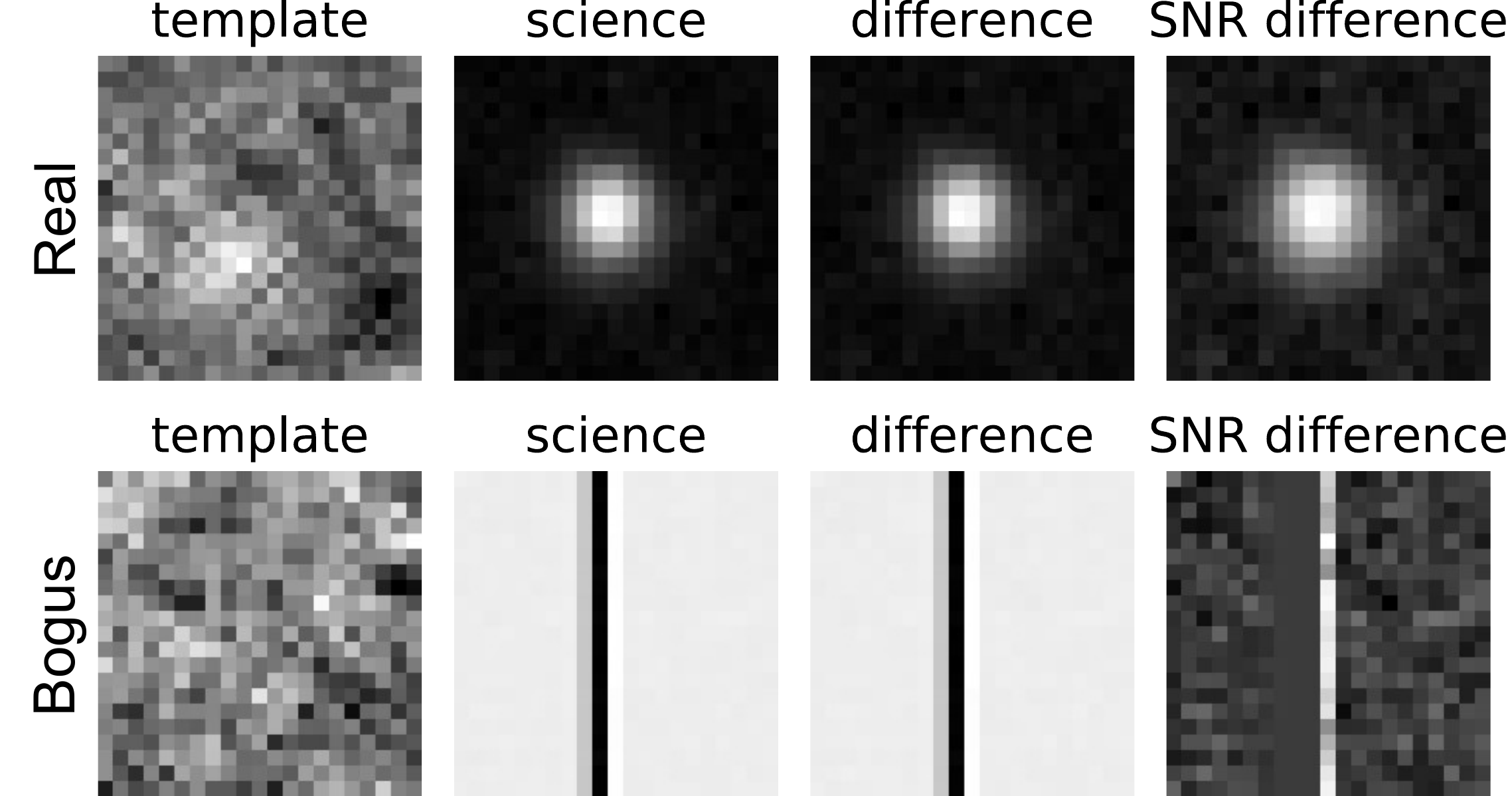}}
\caption{Examples of HiTS real (top) and bogus (bottom) alerts.}
\label{fig:HITS_sample}
\end{figure}

The High Cadence Transient Survey (HiTS) \cite{forster2016high} operated from 2013 to 2015 and aimed at detecting transients in their early stages, mainly SNe \cite{forster2018delay}. The HiTS detection pipeline subtracts reference images from new images coming from the telescope, detects sources and classifies them as real or bogus events. HiTS produced four image stamps of 21$\times$21 pixels centered around each event: \emph{template}, \emph{science}, \emph{difference} and \emph{signal-to-noise ratio} (SNR) \emph{difference}, as shown in Fig. \ref{fig:HITS_sample}.

\begin{figure}[t]
\centerline{\includegraphics[height=0.23\textwidth]{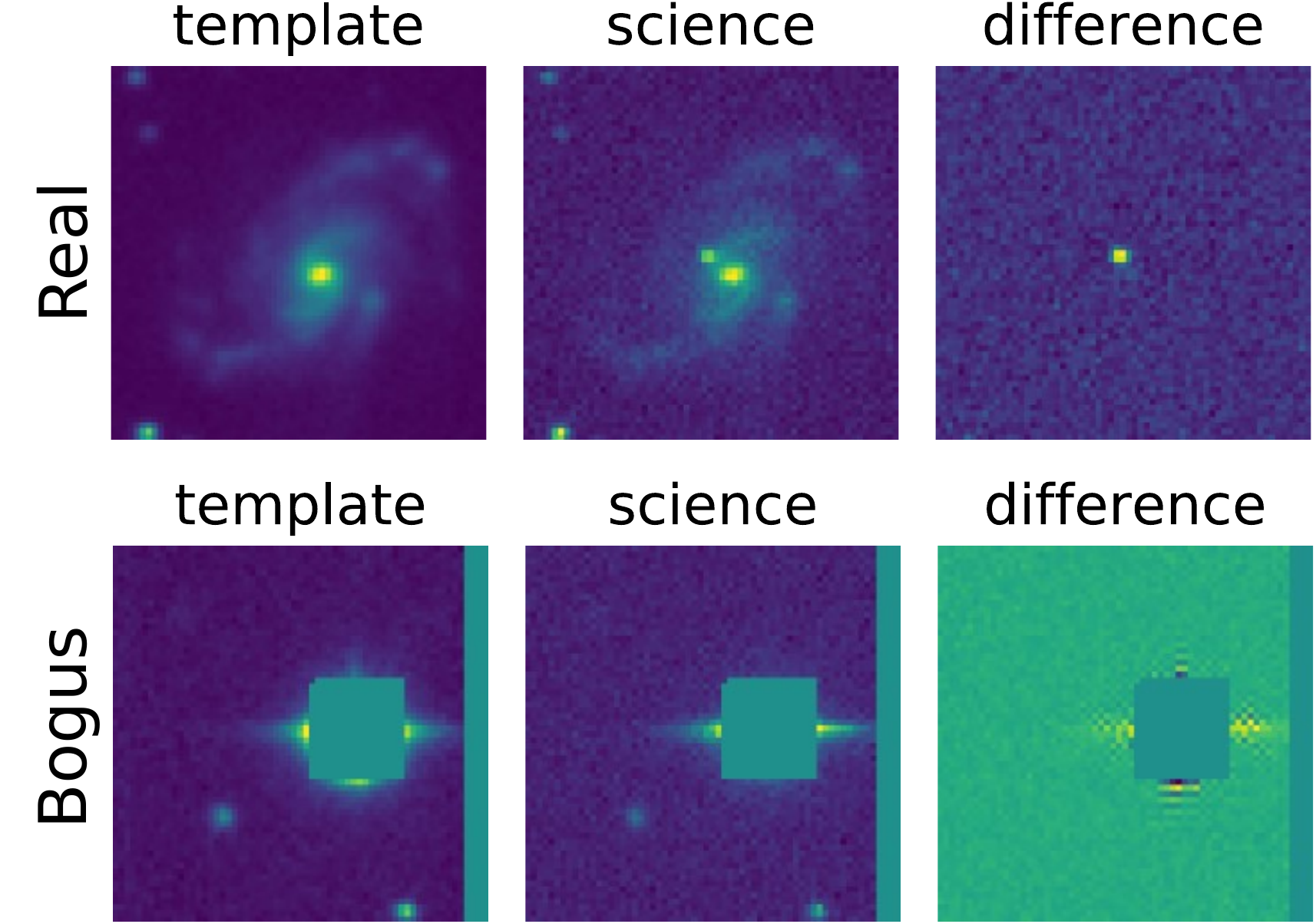}}
\caption{Examples of ZTF real (top) and bogus (bottom) alerts.}
\label{fig:ZTF_sample}
\end{figure}

In HiTS, a discriminative approach was used to distinguish between real and bogus events. As SNe are rare events, to have a balanced dataset, 802,087 SN-like objects were artificially created by injecting bright sources into sky images. 
In \cite{cabrera2016supernovae}, we proposed a Convolutional Neural Network (CNN) to classify sources detected by HiTS as true transients (real) or artifacts (bogus). In 2017 we proposed Deep-HiTS  \cite{cabrera2017deep}, a CNN based model that includes partial rotational invariances. In 2018, we proposed Enhanced Deep-HiTS \cite{reyes2018enhanced} which introduced total rotational invariances. Notice that the aforementioned methods are fully-supervised, so they are not directly comparable with anomaly detection algorithms.

\subsection{Zwicky Transient Facility Survey} 

The Zwicky Transient Facility is a time-domain survey currently in operation, that aims at extending our knowledge of the temporal and dynamic sky.  The real-time pipeline of ZTF generates a stream of image stamps of 63$\times$63 pixels centered around each event, similar to those of HiTS, as shown in Fig.~\ref{fig:ZTF_sample}. ZTF generates its difference images through an image-differencing algorithm \cite{masci2018zwicky}, where the detection of point-source transient events is optimized. ZTF alerts are produced by using a machine-learned classifier along with contextual information. ZTF is a last generation survey that is currently producing a stream of $\sim$1 million alerts per night. Automated systems that are able to ingest, organize and re-distribute all the data from the alert stream are currently being developed.

\subsection{Anomaly Detection and GeoTransform}

Anomaly or outlier detection is a widely studied problem, and there are numerous literature reviews on this topic \cite{aggarwal2015outlier}, \cite{wang2019progress}. Most anomaly detection algorithms work well in the feature space, e.g., IF and One-Class Support Vector Machine (OC-SVM) \cite{scholkopf2001estimating}, but fail to detect anomalies in high dimensional manifolds such as images. To address this issue, many deep learning approaches have been proposed which mainly rely on reconstruction methods such as AutoEncoders (AE) \cite{hinton2011transforming}, data distribution learning such as Generative Adversarial Networks (GANs) \cite{goodfellow2014generative}  and one-class classifiers. 

Recently, algorithms based on transformations over images have been proposed, the first one is GeoTransform \cite{golan2018deep}, where a series of geometric transformations are applied to inlier images. This allows creating a self-labeled dataset, where each transformation has its own label and a classifier is used to discriminate among the transformations made to each sample. It is expected that when an anomaly is presented to the classifier, it would not be able to identify correctly the transformations made. Another transformation based method is Inverse-Transform AutoEncoder \cite{huang2019inverse}, which uses an AE to reconstruct images from their transformed versions. 

\begin{figure*}[t]
\centering
\begin{subfigure}{.45\textwidth}
  \centering
  \includegraphics[width=1\linewidth]{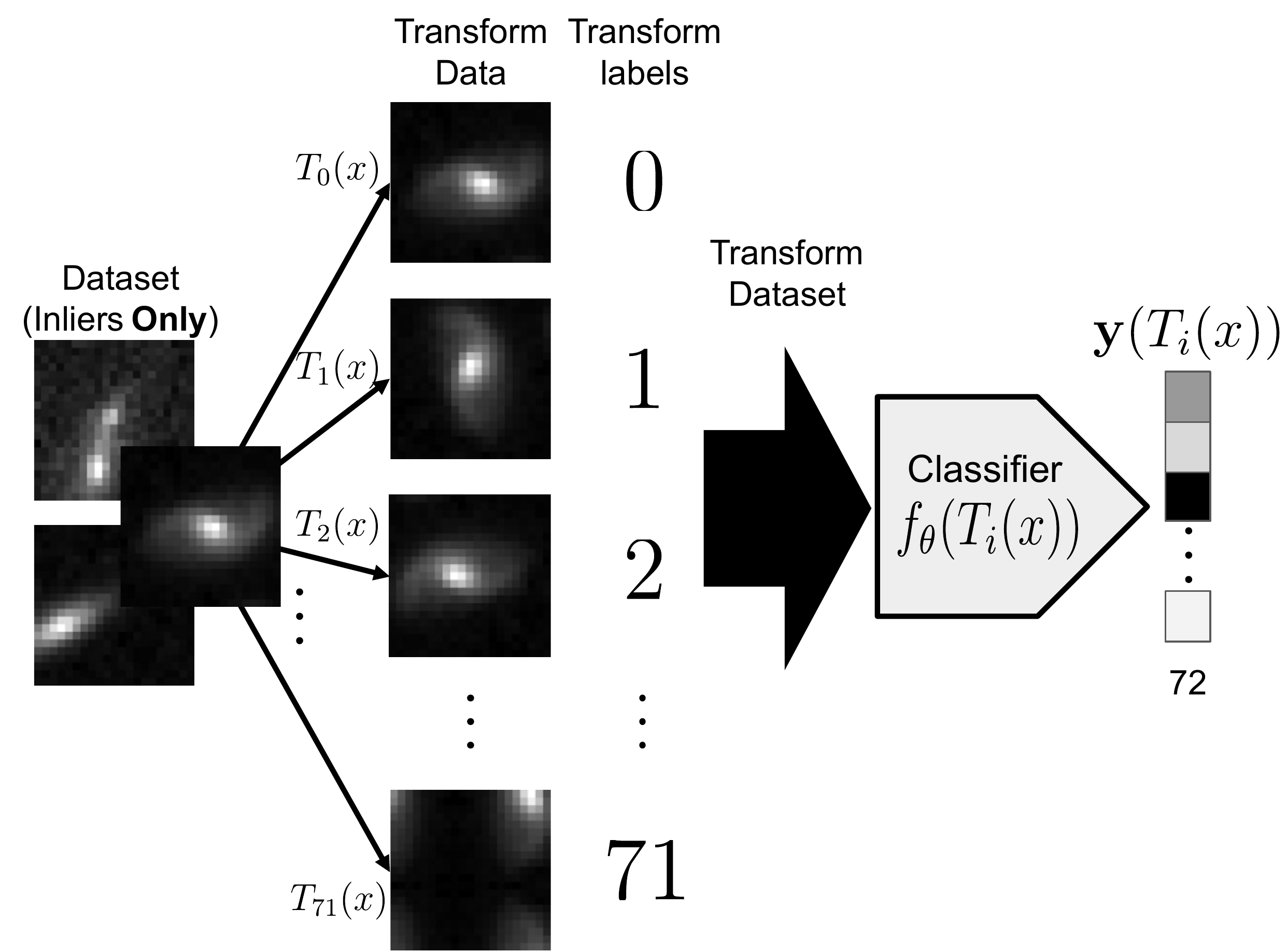}
  \caption{}
  \label{fig:model_train_phase}
\end{subfigure}%
\hspace{0.04\textwidth}
\begin{subfigure}{.5\textwidth}
  \centering
  \includegraphics[width=1\linewidth]{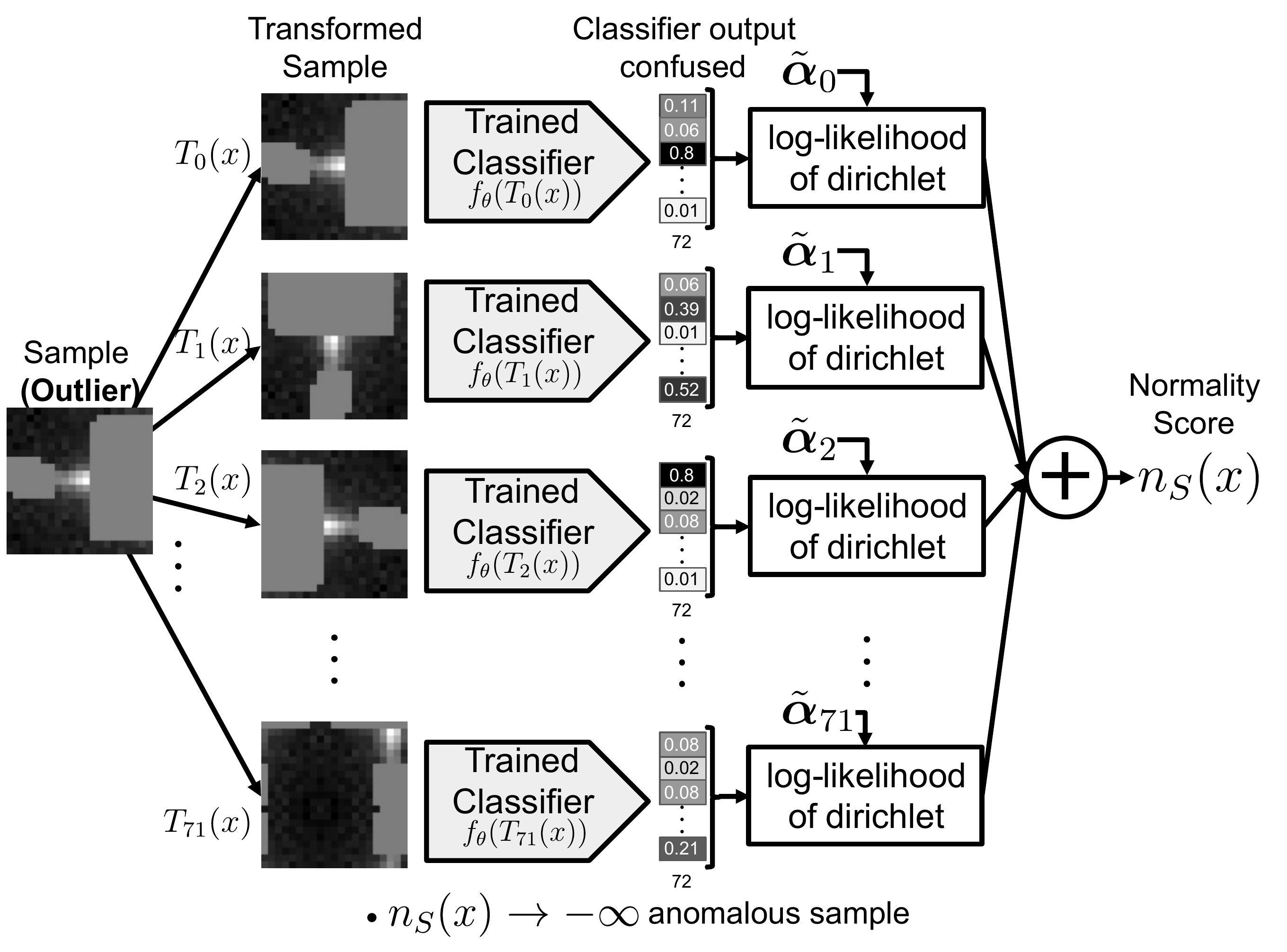}
  \caption{}
  \label{fig:model_test_phase}
\end{subfigure}
\caption{Schematic of GeoTransform inner workings. (a) Training phase of the algorithm, it consists of selecting a set of transformations $\mathcal{T} = \{T_0, T_1, ..., T_{k-1}\}$, for $k=72$, and apply them to the data samples in order to generate a self-labeled dataset $S_\mathcal{T}$, where the labels are the indexes of the transformations. Then a classifier $f_{\theta}$ is trained over the self-labeled dataset, which is used to estimate the parameter $\tilde{\boldsymbol{\alpha}}_i$ of a Dirichlet distribution $Dir(\boldsymbol{\alpha}_i)$ associated with the classifier's output for all the transformed training samples of $S_{\{T_i\}}$, for a given transformation $T_i$. (b) Testing phase of GeoTransform, it is used to evaluate a new sample $x$ by applying all transformations to it, and then getting the classifier's output. For every output, the log-likelihood of the sample is calculated by using the respective transformation's Dirichlet parameter $\tilde{\boldsymbol{\alpha}}_i$, then all the log-likelihoods are summed up to yield the normality score $n_S(x)$. The more negative the score, the more anomalous the sample. In (b) the sample is an outlier and the normality score is low since the classifier cannot discriminate among applied transformations.}
\label{fig:GeoTransfor_model}
\end{figure*}

GeoTransform aims at learning a scoring function that tells how normal (inlier-like) is a given sample $x$, which is defined as the set of images that compose each sample. Fig.~\ref{fig:GeoTransfor_model} shows the inner-workings of the model at the training and testing phases. A self-labeled dataset is generated, where each class corresponds to a specific geometric transformation $T_i$ applied to all the inliers. Transformations are formed by a composition of 9 types of shifts, 4 types of rotations, and flips of the images in each sample, yielding a final set of 72 transformations $\mathcal{T} = \{T_0, T_1, ..., T_{k-1}\}$, where $k=72$. With this self-labeled dataset a $k$-class image classifier $f_{\theta}$, with parameters $\theta$,\textbf{} is trained by using cross-entropy loss,  to correctly estimate the index of the applied transformation $T_i$  for every training sample. 
To evaluate a sample $x$, GeoTransform uses a \emph{Dirichlet normality score} $n_S(x)$, which is defined as the combination of the log-likelihood of the output softmax vectors $\mathbf{y}(x)\triangleq softmax(f_{\theta}(x))$ coming from the classifier $f_{\theta}$, for each transformed sample $T_i(x)$, conditioned on the $i$-th applied transformation $T_i$ as follows:

\begin{equation}
 n_S(x)\triangleq\frac{1}{k}\sum^{k-1}_{i=0}\log p(\mathbf{y}(T(x))|T=T_i). 
 \label{eq:normality_score}
\end{equation}
 
Assuming that all the conditional distributions are independent and follow a Dirichlet distribution, then $\mathbf{y}(T(x))|T=T_i\sim Dir(\boldsymbol{\alpha}_i)$, where $\boldsymbol{\alpha}_i \in \mathbb{R}^k_+$, $x\sim p_X(x)$, $i\sim Uni(0,k-1)$, and $p_X(x)$ is the data distribution of inlier samples. The maximum likelihood parameters  $\boldsymbol{\alpha}_i$, of the Dirichlet distribution $Dir(\boldsymbol{\alpha}_i)$ for every transformation $T_i$, need to be calculated, but as this problem is intractable they are estimated through numerical methods \cite{wicker2008maximum}, \cite{minka2000estimating}. The estimation is expressed as $\tilde{\boldsymbol{\alpha}}_i$, and the final normality score used is the following:

\begin{equation}
 n_S(x)=\frac{1}{k}\sum^{k-1}_{i=0}(\tilde{\boldsymbol{\alpha}}_i -1) \cdot \log \mathbf{y}(T_i(x)). 
 \label{eq:diri_normality_score}
\end{equation}

The aforementioned process is the training phase of the GeoTransform algorithm, which is illustrated in Fig. \ref{fig:model_train_phase}. To compute the normality score $n_S(x)$ over a new sample, all the transformations are applied to it and then the softmax outputs of the classifier are calculated. For every softmax vector $\mathbf{y}(T_i(x))$, the log-likelihood is computed using the respective  $\tilde{\boldsymbol{\alpha}}_i$ fitted with the training data, and finally all the log-likelihoods are added to yield $n_S(x)$. This procedure is illustrated in Fig. \ref{fig:model_test_phase}.

\section{Problem Statement}

In this work, we use the terms anomaly and outlier interchangeably. We consider an anomaly as any sample that is significantly different from the inliers. At training time the algorithms used for anomaly detection have access to the inlier dataset only. At test time, a balanced set where half of the elements are outliers and the other half are inliers is used. The goal is to learn a binary classifier that outputs 1 for inlier samples and 0 for outlier samples. Usually, a normality score $n_S(x): \mathcal{X}\rightarrow\mathbb{R}$ that maps the space of all possible images $\mathcal{X}$ to a scalar value is proposed. The higher the normality score the more inlier-like the sample $x$. In this work, we first focus on getting the best possible normality scoring function $n_S(x)$ in terms of the area under the Receiving Operating Characteristic (ROC) curve, denoted as AUROC. For evaluating the accuracy over the astronomical datasets, we define a threshold $\lambda$ to classify samples as inliers or outliers, reporting the accuracy on balanced datasets.

\section{Transformation Modifications}

The GeoTransform method \cite{golan2018deep} uses a user-defined set of transformations $\mathcal{T} = \{T_0, T_1,...T_{k-1}\}$,  where each geometric transformation $T_i: 1\leq i\leq k-1$ changes the values of pixels of the original images, and $T_0(x)=x$ is the identity transformation. Given $\mathcal{T}$, a self-labeled dataset $S_{\mathcal{T}}$ is built, where a label is associated with each sample corresponding to the index of the transformation applied to it, i.e. for any $x \in S$, where $S$ is the original non transformed dataset, the label for the transformed sample $T_i(x)$ is $i$. In the following subsections, we describe our proposed enhancements to the GeoTransform method for anomaly detection.

\subsection{New Transformations}
\label{subsec:new_transforms}

The authors of GeoTransform claim that geometric transformations get better results over non-geometric ones because the former preserves the spatial information and local pixel correlation of inlier-like images. They tried non-geometric transformations on MNIST and CIFAR10 such as sharpening, Gaussian blurring and gamma correction, but obtained a deteriorated performance. Therefore, they discarded them with the hypothesis that non-geometric transformations reduce performance because they erase features from $S_{\mathcal{T}}$ that are important to characterize inlier images.

\begin{figure}[!t]
\centerline{\includegraphics[width=0.348\textwidth]{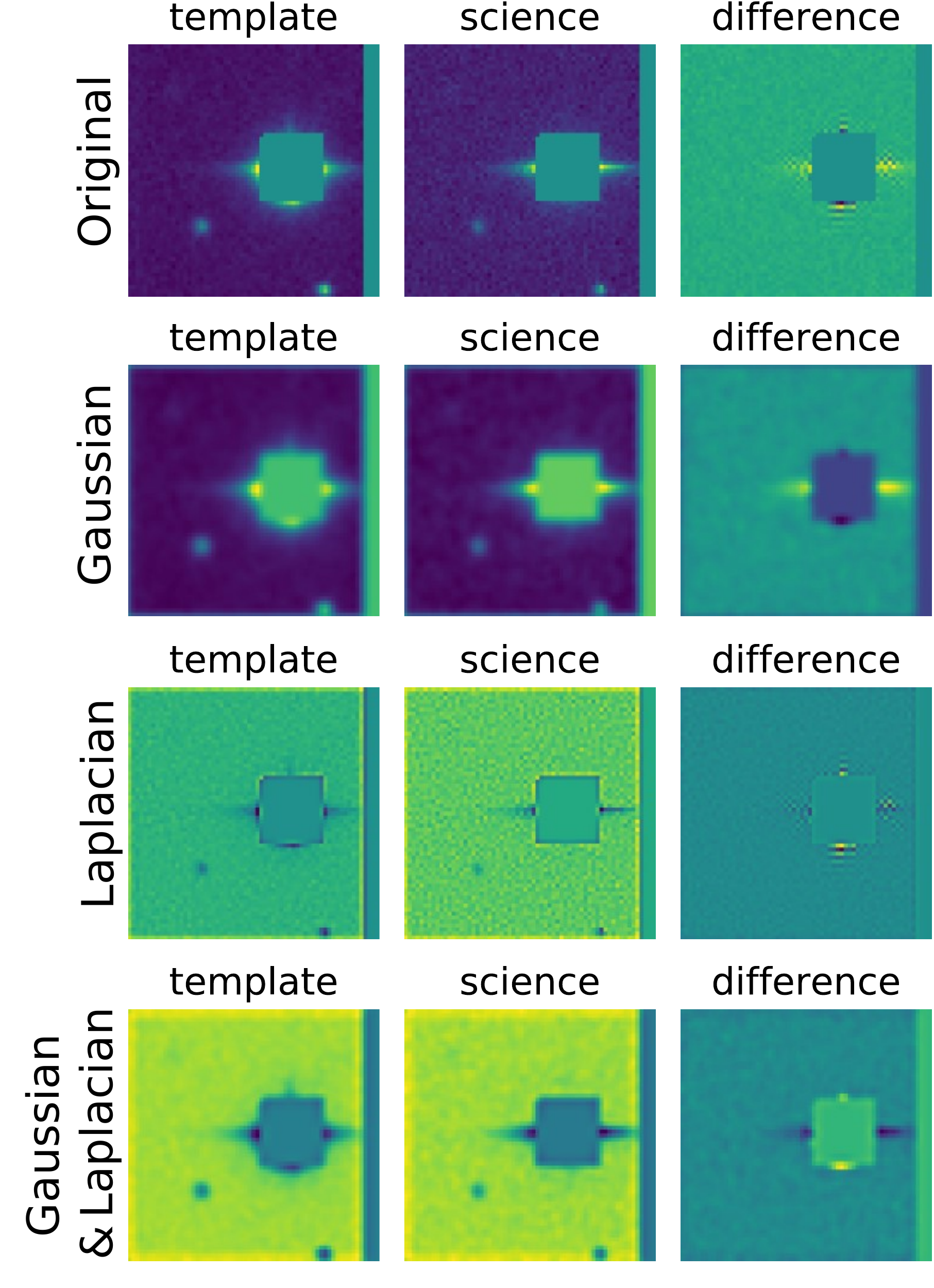}}
\caption{Effect of applying Gaussian and Laplacian filter based transformations to a bogus ZTF sample.}
\label{fig:ZTF_new_transforms}
\end{figure}

Our hypothesis is that finding the right set of transformations is problem dependent and that we can design new features using a-priori knowledge. Astronomical artifacts usually appear in difference images as sharp edges that significantly differ in value with respect to the background sky. We postulate that adding edge detection or edge erasing transformations to GeoTransform will be useful to highlight features present only in some astronomical artifacts. To encode this a-priori knowledge of astronomical artifacts, we use transformations based on \emph{Laplacian} and \emph{Gaussian} filters, in order to perform edge detection and edge blurring/erasing \cite{gonzalez2008digital}, respectively. Fig. \ref{fig:ZTF_new_transforms} shows how the proposed transformations affect a bogus sample, when Gaussian filtering is applied the edges are blurred, whereas for the Laplacian filter edges are highlighted, and when combining both of them, with the Gaussian filter applied first, only edges not completely blurred by the Gaussian filter are highlighted by the Laplacian filter. For both filters a kernel size of 5$\times$5 is used, for the Gaussian kernel we used $\sigma=1$ and for the Laplacian kernel $\sigma=0.5$.

\subsection{Transformation Selection}

\label{subsec:Transform_select}

The original set of 72 transformations of GeoTransform may work well for datasets without geometric invariances, a property that we usually do not know a-priori. Herein, we propose a novel strategy to discard useless transformations, i.e., those that are redundant can be excluded. Our strategy is to use a neural network classifier to discriminate among pairs of transformations. To do so, we take the $S_{\mathcal{T}}$ self-labeled dataset with $|\mathcal{T}|$ transformations and split it into binary subsets $S_{\mathcal{T}_{ij}}$ composed of a pair of transformations $T_i$ and $T_j$, where $i\neq j$ and $i>j$. Thus $\frac{|\mathcal{T}| \times (|\mathcal{T}|-1)}{2}$ subsets are obtained. For each subset $S_{\mathcal{T}_{ij}}$ we train a classifier using the Deep-HiTS architecture \cite{cabrera2017deep}, without rotated images at the input.

If the dataset is invariant to a given transformation, the classifier won't be able to tell apart one transformation from the other and the accuracy would be around~50\%. Computing the accuracy for every pair of transformations, we get a \emph{discrimination matrix} which tells us if a given pair of transformations can be discriminated by a classifier or not. 
After computing the discrimination matrix, for every pair of transformations that get an accuracy between 49\% and 51\%, we select the transformation with the least number of operations and discard the other one. The rationale is that it doesn't matter which transformation should be chosen from the pair because the classifier cannot distinguish between them.

Applying the aforementioned procedure to the set of transformations used by GeoTransform, we can identify which ones are useless due to the presence of geometric invariances within a dataset. Eliminating redundant transformations has the benefit of using fewer resources and time to calculate extra transformations, making the algorithm faster, and reducing the complexity of the classification space.

\section{Experiments}

In the following subsections, we describe the baseline methods used to compare GeoTransform and the proposed enhancements, as well as the astronomical datasets used in this work and the experimental setup. During the training phase, all methods have access to the inlier class only. Performance is measured in terms of AUROC over balanced test sets containing both inliers and outliers. The results shown in every table are the mean and standard deviation of 10 runs.

\subsection{Baseline Methods}

\textbf{One-Class Support Vector Machine.} OC-SVM is a kernel based method for anomaly detection. It transforms data from feature space to a Reproducing Kernel Hilbert Space where it learns a decision boundary trying to enclose all inlier samples in a compact space, and everything that lies outside the boundary it is presumed anomalous. We use OC-SVM in two different ways, the first one converts images to a single flattened vector, and the second one uses as inputs the bottleneck representation of a convolutional AE, with the same setup as in \cite{golan2018deep}. We call these two approaches RAW-OC-SVM and CAE-OC-SVM, respectively. We gave an advantage to OC-SVM in both setups, since we chose the hyperparameters $\gamma$ and $\nu$ that got the best AUROC over the test set using a grid search: $\gamma \in \{2^{-7}, 2^{-6}, ..., 2^{2}\}$, $\nu \in \{0.1, 0.2, ..., 0.9\}$. For the convolutional AE architecture in CAE-OC-SVM we adapt the discriminator and the generator of DCGAN \cite{radford2015unsupervised}, to the encoder and decoder of the AE, respectively.

\textbf{Isolation Forest.} IF is a tree based algorithm that works by isolating each sample with boundaries made by decision trees. The fewer the decision boundaries needed to isolate a sample the more anomalous it is deemed. In the same way as OC-SVM, hyperparameters of IF were chosen to maximize AUROC over the test set using a grid search: $n\_estimators \in \{100, 200, ..., 800\}$, $contamination \in \{0.1, 0.2, ..., 0.5\}$.

\textbf{Deep structured energy-based models.} It is a deep learning based model with the acronym DSEBM \cite{zhai2016deep}, it works by yielding the negative log probability (energy function) associated with an input. When a sample has a high energy value it is considered to be anomalous. The architecture for this model is the same as the encoder for CAE-OC-SVM.

\textbf{Anomaly Detection GAN.} ADGAN \cite{deecke2018image} uses a Generative Adversarial Network (GAN) to learn the mapping of a multivariate Gaussian distribution (latent space), to the inlier training dataset distribution. At evaluation time, to compute the outlier score of a new sample $x$, the Mean Square Error (MSE) is calculated between $x$ and a GAN generated sample $G(z)$, where $z$ is a vector coming from the latent space. Then $z$ is modified through gradient descent to minimize the MSE, and this process is iteratively performed 5 times. The anomaly score is defined as the MSE between the original sample $x$ and the last generated sample $G(z_5)$. The architectures of the generator and discriminator are the same as the decoder and encoder of CAE-OC-SVM, respectively.

\textbf{Multiple-Objective Generative Adversarial Active Learning.} MO-GAAL \cite{liu2019generative} is a state-of-the-art GAN based method that tries to learn multiple sub-optimal GANs trained on inliers in order to generate outliers. It tries to populate the space around inliers with the generated outliers and then it trains a classifier to distinguish between inliers and the GAN generated outliers, expecting that the actual outliers will fall in the same class as the GAN generated outliers. 

\begin{figure}[t]
\centerline{\includegraphics[width=0.36 \textwidth]{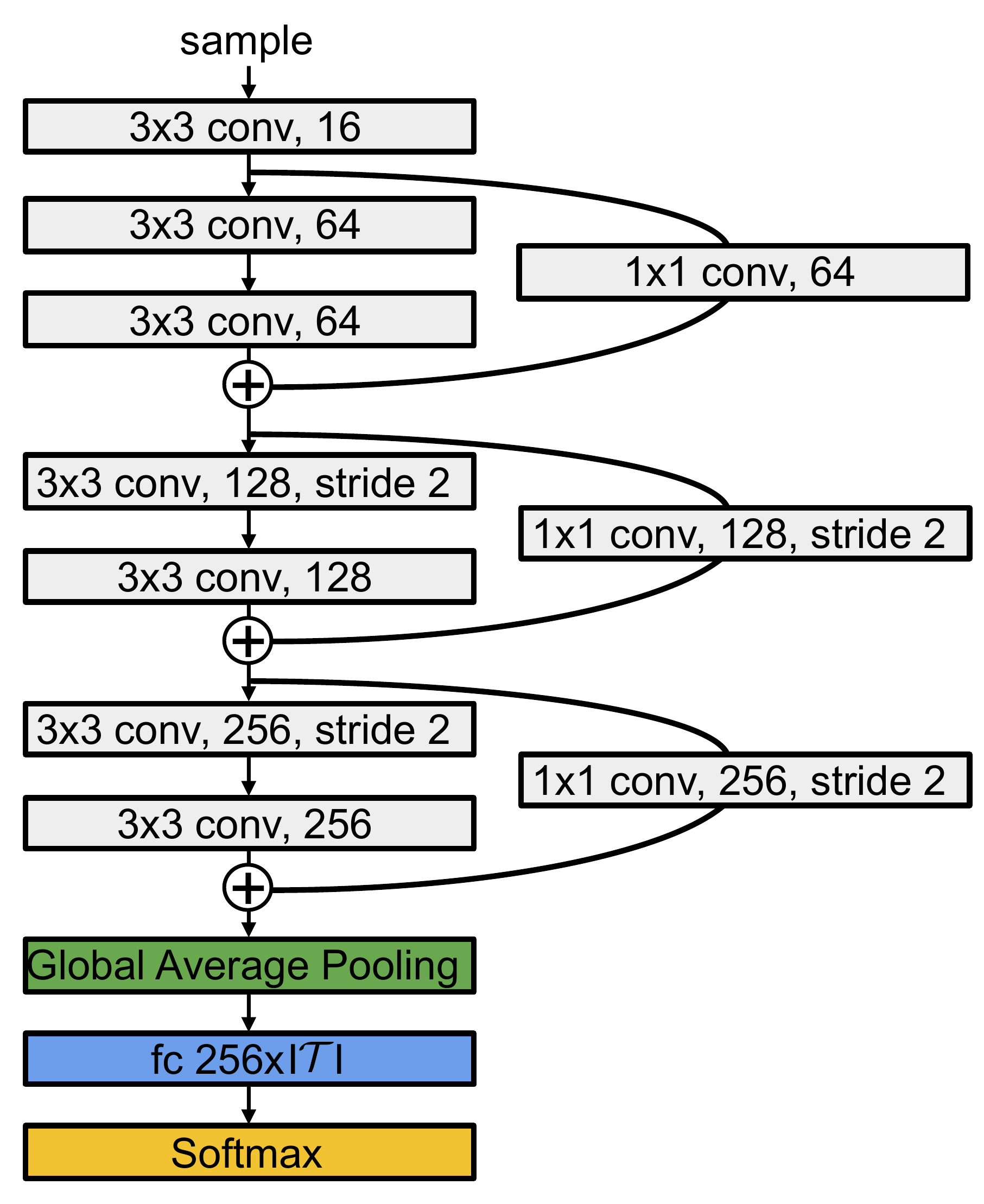}}
\caption{ Wide Residual Network (WRN) architecture of depth 10 and widen factor 4, including 7 convolutional layers, 3 skip connections, a global average pooling and a fully connected layer with output size equal to the number of applied transformations $|\mathcal{T}|$. At the beginning and end of every skip connection, batch normalization with momentum 0.9 and $\epsilon=1^{-5}$ is applied.}
\label{fig:wide_resnet_architecture}
\end{figure}

\textbf{GeoTransform.} It uses a Wide Residual Network (WRN) \cite{zagoruyko2016wide} with architecture parameters of depth 10 and widen factor 4. The full architecture consists of 7 convolutional layers and 3 residual connections, as shown in Fig. \ref{fig:wide_resnet_architecture}. A batch size of 128, ADAM optimizer and a cross-entropy loss are used. In \cite{golan2018deep} they train the network for $[200/\mathcal{T}]$ epochs over the self-labeled dataset $S_{\mathcal{T}}$, to mimic the same number of iterations as if the WRN was trained on $S$ for 200 epochs. The original model uses a set of $|\mathcal{T}|=72$ transformations. As this configuration will vary in different experiments, we will refer to it as GeoTransform72, and new models will be denoted as GeoTransform$Number\_of\_transformations$.

\subsection{Astronomical Datasets}
\textbf{HiTS.} It contains 1,604,174 labeled samples, where half of them are bogus and the other half are real astronomical alerts. This survey was elaborated with the aim of identifying supernovae (SNe). Each sample of HiTS has four 21$\times$21 images that were individually normalized to have pixel values in the range $[-1,1]$.  For all experiments, we divided the data into 3 sets, a training set with 7,000 samples composed by inliers only, a validation set with 1,000 inlier samples, and a balanced test set of 2,000 inliers and 2,000 outliers.

\textbf{ZTF.} This is a custom made dataset built by the ALeRCE team, based on the ZTF stream of alerts. It consists of $\sim$35,000 samples labeled as real alerts and 5,321 samples labeled as bogus. The process of labeling real alerts was made through cross-matching \cite{zhao2009paralleled} and includes classes such as variable stars (VS), active galactic nuclei (AGN), asteroids, and supernovae (SNe). On the other hand, experts labeled bogus examples by hand. Each ZTF sample consists of 3 images that are cropped at the center to 21$\times$21 pixels and then are individually normalized to have pixel values in the range $[-1,1]$.  There are images with NaNs in ZTF because some of the light sources were erased to prevent pixel value saturation within an image. NaN values were replaced by 0's. For the anomaly detection experiments we divided the data into 3 sets, a training set with 7,000 samples composed only by inliers, a validation set with 1,000 inlier samples, and a balanced test set of 3,000 inliers and 3,000 outliers.

\subsection{Modifying the Training Process}

In \cite{golan2018deep} a fixed number of training epochs are used to follow the same setup as the baseline deep learning methods. The latter uses 200 training epochs, but because GeoTransform operates with a self-labeled dataset $S_{\mathcal{T}}$ that is $\mathcal{T}$ times bigger than the original one, they adjust the training process to perform the same number of iterations as the baseline methods, i.e., $200/\mathcal{T}$ epochs. There is a direct correlation on how well the transformations are classified and the normality score. However, in the training process, nothing prevents the model from overfitting to the training set. We use a validation set composed of inliers, enabling us to adopt an early-stopping criterion. For each run, we perform training with early-stopping with patience 0 and check the validation loss at the end of each epoch. Patience 0 means that whenever the validation loss increases, training is stopped. Table \ref{table:sota_val} shows the AUROC obtained when using training for $200/\mathcal{T}$ and the proposed method of early-stopping. In both datasets, the performance of GeoTransform72 improved when using early-stopping. In what follows, all trained GeoTransform models include early-stopping, with exception of the \emph{original GeoTransform72}.

\subsection{Improving Anomaly Detection with New Transformations}

As mentioned in Subsection \ref{subsec:new_transforms} we propose including Gaussian and Laplace kernel transformations to exploit a-priori knowledge of astronomical artifacts. As a kernel operation can be applied or not, this adds two possible states to the composition of transformations, e.g. as the original GeoTransform has 72 transformations, including a new one means that we double the number of transformations to 144, in order to include every possible combination of the current pool. To avoid the computational cost of generating too many transformations, we use a simplified setup by performing the kernel transformation \emph{only} over shift operations, which have 9 possible states. In this case, including a new kernel operator adds 9 operations only, so for the same example above, if we add a kernel transformation to the composition that generates the original 72 transformations, they increase to 81. Following this procedure, we compare the original GeoTransform72, with the simplified setup of GeoTransform adding only Gaussian kernels (GeoTransform81-G) or only Laplacian kernels (GeoTransform81-L) and using both kernel operations with 99 transformations (GeoTransform99). For the simplified setup of 99 transformations, 9 transformations are Gaussian kernels, 9 Laplacian kernels and 9 combinations of both types of kernels. Applying either Gaussian or Laplacian kernels over the whole set of 72 transformations we get 144 transformations, i.e. GeoTransform144-G for Gaussian kernels and GeoTransform144-L for Laplacian kernels, while applying both kernels yields 288 transformations (GeoTransform288).  When both kernel operations are present we first apply the Gaussian kernel followed by the Laplacian kernel.

In Table \ref{table:new_transforms} we show the AUROC results of applying the proposed transformations over HiTS and ZTF datasets. It can be observed that the addition of any of the kernel transformations improves the baseline results for both datasets. The best results are achieved when both the Gaussian and Laplace kernels are used over shift transformations in the setup with 99 transformations (GeoTransform99). For both datasets, an AUROC improvement is obtained using  GeoTransform99 with respect to GeoTransform72, and the Welch's hypothesis test shows p-values of less than 2.9$\times$10$^{-3}$ for HiTS dataset and less than 4.0$\times$10$^{-6}$ for ZTF dataset, both p-values are statistically significant.

\begin{table}[t]
\centering
\normalsize
\caption{Alternative training scheduling methods for GeoTransform. This table compares the original scheduling of training during $200/\mathcal{T}$ epochs, versus the usage of an early-stopping criterion.}
\label{table:sota_val}
\resizebox{\columnwidth}{!}{%
\begin{tabular}{c|c|c|c}
\hline
\begin{tabular}[c]{@{}c@{}}Model\end{tabular}   &
\begin{tabular}[c]{@{}c@{}}Training  \\ scheduling\end{tabular}   & \begin{tabular}[c]{@{}c@{}}HiTS\\  AUROC\end{tabular} & \begin{tabular}[c]{@{}c@{}}ZTF\\ AUROC\end{tabular} \\ \hline \hline
\begin{tabular}[c]{@{}c@{}}Original  \\ GeoTransform72 \cite{golan2018deep}\end{tabular}      & 
 $200/\mathcal{T}$ epochs      &  98.60$\pm$0.23                                         & 85.63$\pm$1.48                                        \\ \hline
 GeoTransform72 &
\begin{tabular}[c]{@{}c@{}}Early-stopping \\  with patience 0\end{tabular} & \textbf{98.78$\pm$0.26}                                 & \textbf{87.33$\pm$1.24}                               \\ \hline
\end{tabular}%
}
\end{table}

\begin{table}[t]
\centering
\normalsize
\caption{Effect of including different kernel based transformations in the original pool of 72 transformations. }
\label{table:new_transforms}
\resizebox{\columnwidth}{!}{%
\begin{tabular}{c|c|c|c}
\hline
\begin{tabular}[c]{@{}c@{}}Model\end{tabular}   &
\begin{tabular}[c]{@{}c@{}}Transformations setup \\ (\# transformations)\end{tabular}   & \begin{tabular}[c]{@{}c@{}}HiTS\\  AUROC\end{tabular} & \begin{tabular}[c]{@{}c@{}}ZTF\\ AUROC\end{tabular} \\ \hline \hline
GeoTransform72 & Original (72)      & 98.78$\pm$0.26                                          & 87.33$\pm$1.24                                        \\ \hline
GeoTransform81-G &
\begin{tabular}[c]{@{}c@{}}Original +\\ Gauss on Shifts (81)\end{tabular}     & 99.01$\pm$0.07                                                       &     89.28$\pm$1.30                                                \\ \hline
GeoTransform81-L &\begin{tabular}[c]{@{}c@{}}Original +\\ Laplace on Shifts (81)\end{tabular}        &     98.87$\pm$0.11                                                    &  88.94$\pm$0.96                                                   \\ \hline
GeoTransform99 &\begin{tabular}[c]{@{}c@{}}Original + (Gauss \&\\  Laplace) on Shifts (99)\end{tabular} & \textbf{99.12$\pm$0.04}                                 & \textbf{90.80$\pm$0.61}                               \\ \hline
GeoTransform144-G &\begin{tabular}[c]{@{}c@{}}Original + Gauss \\  (144)\end{tabular} &  98.84$\pm$0.15                                                                   &  87.58$\pm$0.84                             \\ \hline
GeoTransform144-L &\begin{tabular}[c]{@{}c@{}}Original + Laplace \\  (144)\end{tabular}&               98.85$\pm$0.15                                                    &  87.62$\pm$1.00                          \\ \hline
GeoTransform288 &\begin{tabular}[c]{@{}c@{}}Original + (Gauss \&\\  Laplace) (288)\end{tabular} &    99.02$\pm$0.05                                                   &  89.79$\pm$0.67                                  \\ \hline\hline
\multicolumn{2}{c|}{Welch's t-test p-value (72) v/s (99)} &    2.9$\times$10$^{-3}$                                                   &  4.0$\times$10$^{-6}$                                  \\ \hline
\end{tabular}
}
\end{table}

\begin{table*}[t!]
\centering
\normalsize
\caption{Performance of the transformation selection strategy over GeoTransform72 and GeoTransform99. Applying transformation selection to the set of transformations of GeoTransform72, reduces it to GeoTransform9 for both datasets. In the case of ZTF, GeoTransform9 obtained a lower AUROC but higher accuracy than those obtained with  GeoTransform72. Applying the selection strategy to GeoTransform99 reduces the number of transformations to GeoTransform35 for HiTS and GeoTransform29 for ZTF. The latter two models achieved the best performance in terms of AUROC.}
\label{table:transform_selection}
\resizebox{\textwidth}{!}{%
\begin{tabular}{c|c|c|c|c|c}
\hline
\multirow{2}{*}{Model} & Transformation & \multicolumn{2}{c|}{HiTS}                  & \multicolumn{2}{c}{ZTF}          \\ \cline{3-6} 
                                                                                               &  selection setup     & AUROC                 & Accuracy & AUROC        & Accuracy  \\ \hline \hline
GeoTransform72 & None                  & 98.78$\pm$0.26          & 96.97$\pm$0.55                   & 87.33$\pm$1.24 & 77.80$\pm$0.94                   \\ \hline
GeoTransform9 & Transformation selection over GeoTransform72                               & 99.16$\pm$0.13          &    97.46$\pm$0.22                & 86.49$\pm$1.61 & 78.11$\pm$1.33                   \\ \hline
GeoTransform99 & None                                                              & 99.12$\pm$0.04          & 97.59$\pm$0.19          & 90.80$\pm$0.61 &  83.53$\pm$0.66       \\ \hline
\begin{tabular}[c]{@{}c@{}}(HiTS) GeoTransform35\\(ZTF) GeoTransform29\end{tabular} & Transformation selection over GeoTransform99                                                                  & \textbf{99.20$\pm$0.06}          & 97.21$\pm$0.10                   & \textbf{91.39$\pm$0.76} &     82.81$\pm$0.36               \\ \hline\hline
\multicolumn{2}{c|}{\begin{tabular}[c]{@{}c@{}c@{}}Welch's t-test p-value GeoTransform72 v/s \\ GeoTransform35 \& GeoTransform29\end{tabular}} &    6.8$\times$10$^{-4}$  &       2.1$\times$10$^{-1}$                                            &  5.3$\times$10$^{-7}$ &     1.1$\times$10$^{-8}$                             \\ \hline
\end{tabular}}
\end{table*}

\subsection{Transformation Selection Results}

\begin{figure}[t]
\centerline{\includegraphics[width=0.49 \textwidth]{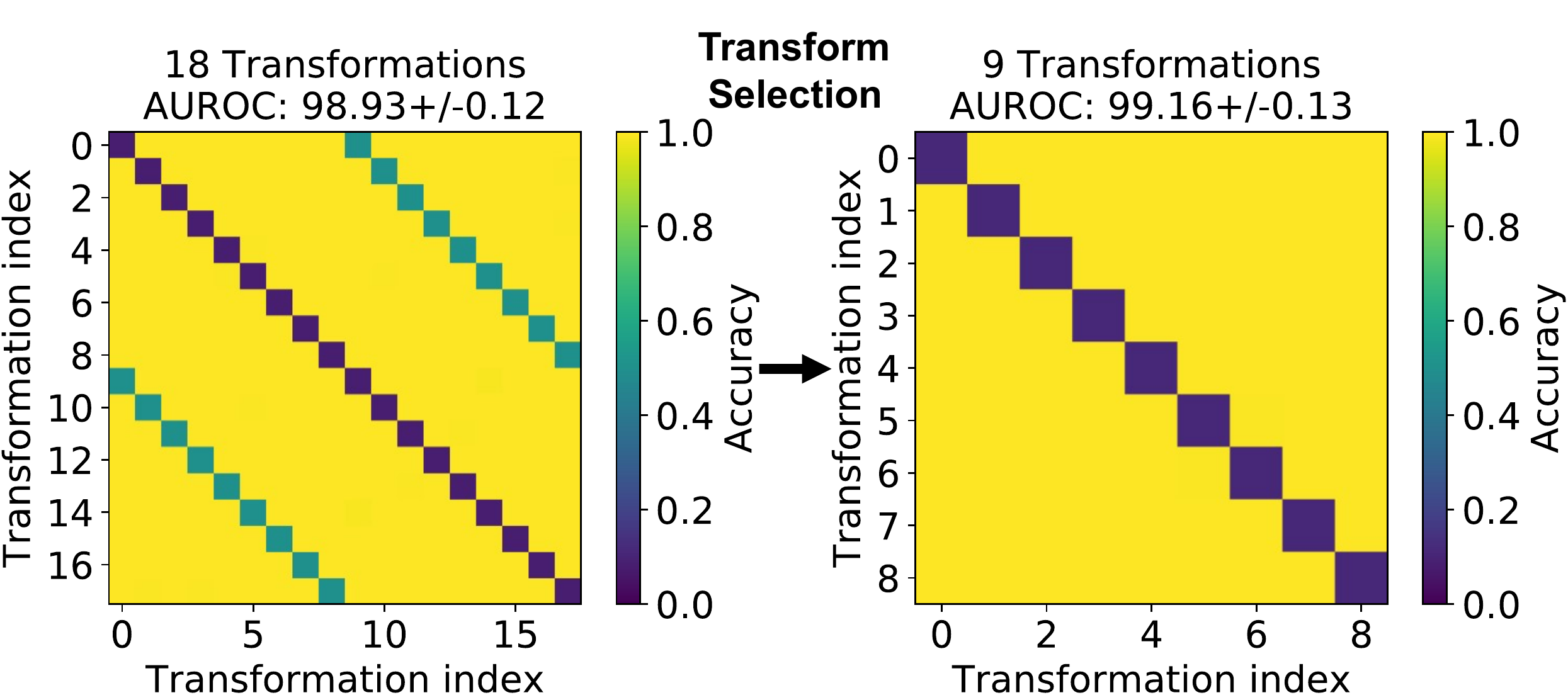}}
\caption{Discrimination matrix for the HiTS dataset, using a toy example of 18 transformations composed of horizontal flip and shifts. Likewise a correlation matrix, the upper triangle and lower triangle of the matrices are identical and the diagonals are meaningless. The matrix shows that shifts can be easily discriminated ($\sim$100\% accuracy), except for the pair of transformations $(T_i,T_{i+9}), i\in\{1,2,...,9\}$ (green squares) that gets an accuracy of $\sim$50\%. A classifier is unable to distinguish between a shifted image that has been previously flipped. On the right side, the discrimination matrix after selecting transformations is plotted, eliminating redundant transformations. In this example, the AUROC improves when selecting 9 transformations out of 18 (eliminating the flip).}
\label{fig:HiTS18to9}
\end{figure}

To filter out useless transformations in the models shown in Table \ref{table:new_transforms}, we use the discriminator matrix described in \ref{subsec:Transform_select}. Those transformations that get a $\sim$50\% accuracy in the discriminator matrix are removed, and then GeoTransform is trained on the remaining transformations. We applied the transformation selection strategy to HiTS, which is presumed to have rotational and flip invariances. We tested this hypothesis with a toy example of flip composed with shift transformations, generating a total of 18 transformations. The classifier should not be able to distinguish between the flipped versions. The discriminator matrix is shown on the left-hand side of Fig.~\ref{fig:HiTS18to9}, where a green square means that the flipped version of a shift cannot be discriminated against.  After eliminating flips, 9 useful transformations remain, corresponding to shifts only, which confirms the aforementioned hypothesis. The right-hand side of Fig. \ref{fig:HiTS18to9} shows the discriminator matrix of the 9 selected transformations, where no redundant transformations remain. An AUROC improvement from 98.93$\pm$0.12 for 18 transformations to 99.16$\pm$0.13 for 9 transformations was obtained. 

 We applied the transformation selection procedure to GeoTransform72 and GeoTransform99 trained on HiTS and ZTF datasets. Table \ref{table:transform_selection} summarizes the results. In addition to the AUROC, we calculate the accuracy for all models. The $\lambda$ threshold for the normality score is set up so that it leaves 97.7\% of the validation inliers on the right side of the threshold, this rule is inspired by the $2\sigma$ interval of a normal distribution. Everything that is higher than $\lambda$ is considered an inlier, otherwise, it is classified as an anomaly. Accuracies for all models are computed using the above procedure.
 
 When applying the transformation selection strategy to GeoTransform72, in both datasets only the 9 shift transformations remained, yielding GeoTransform9. The transformation selection procedure allowed reducing the dimensionality of the problem without sacrificing performance, with the exception of ZTF, where AUROC diminished by 0.84 points, although the accuracy increased by 0.31. For GeoTransform99, we expected that only the 36 transformations of both kernels applied to shifts would be selected, however in the HiTS dataset 35 transformations were selected (GeoTransform35), leaving out the non-shifted Gaussian kernel transformation, and in ZTF 29 transformations were selected (GeoTransform29), removing all but two of the only Gaussian kernel transformations. Table \ref{table:transform_selection} shows that GeoTransform35 for HiTS and GeoTransform29 for ZTF are better than GeoTransfom99, i.e., there is an improvement on the AUROC when applying transformation selection over GeoTransform99 for both datasets. According to Welch's hypothesis test, there are statistically significant differences between GeoTranform72 and the best models of Table \ref{table:transform_selection}, for both datasets.

\subsection{Comparison with Baseline Methods}

Table \ref{table:astro_all_methods} shows a comparison among all baseline methods and the enhanced versions of GeoTransform on the HiTS and ZTF datasets. The enhanced versions of GeoTransform perform better than the original GeoTransform72 without early-stopping, for both datasets. Applying transformation selection over GeoTransform99 yielded the best performing model: GeoTransform35 for HiTS and GeoTransform29 for ZTF. These enhanced models have statistically significant differences when compared to the original GeoTransform72 without early-stopping, according to Welch's hypothesis test.

\section{Conclusions}

In this work, we proposed several enhancements to the GeoTransform algorithm for anomaly detection on astronomical datasets. First, we introduced kernel based transformations such as Laplacian and Gaussian filters to highlight known properties of astronomical artifacts and make them easily distinguishable from real alerts.  In addition, we developed a novel strategy for selecting transformations that allow us to discard useless transformations. Using this procedure GeoTransform72 was reduced to GeoTransform9 on both datasets, and GeoTransform99 was reduced to GeoTransform35 in HiTS and GeoTransform29 in ZTF. Along with the dimensionality reduction, we obtained an improved performance using both AUROC and accuracy measures for all enhanced models. The proposed method gives also an insight into the type of invariances present in a dataset. We compared our best models with several baselines, obtaining the top performing average AUROC of 99.20\% for HiTS and 91.39\% for ZTF. The improvement over the original GeoTransform algorithm is significant both statistically and in practice. Being able to identify bogus events with an automatic method will enable the labeling of bogus class without appealing to human experts. The proposal of new transformations and a selection strategy is a step forward in the improvement of transformation based anomaly detection algorithms. However, the method has room for improvement, e.g., it is worth investigating a way of learning useful transformations. In addition, a clearer theoretical insight on why these kinds of methods work well is needed.


\begin{table}[t]
\centering
\normalsize
\caption{Results of baseline methods and enhanced GeoTransform models on both datasets. The original GeoTransform72 does not include early-stopping.}
\label{table:astro_all_methods}
\resizebox{\columnwidth}{!}{%
\begin{tabular}{c|c|c}
\hline
\multirow{2}{*}{\begin{tabular}[c]{@{}c@{}}Model\end{tabular}} & HiTS                  & ZTF          \\  
                                                                                                      & AUROC                 &  AUROC        \\ \hline \hline
RAW-OC-SVM \cite{scholkopf2001estimating}                       & 97.46$\pm$0.09
                      & 86.01$\pm$0.05                     \\ \hline
CAE-OC-SVM \cite{golan2018deep}                                                                                      & 95.67$\pm$0.14                      &  81.15$\pm$0.20                     \\ \hline
IF \cite{liu2008isolation}                                                                       &            96.10$\pm$0.18                       &       82.19$\pm$0.10                           \\ \hline
DSEBM  \cite{zhai2016deep}                     &          93.52$\pm$1.66 &                           77.69$\pm$0.01                           \\ \hline
ADGAN \cite{deecke2018image}                     & 91.83$\pm$0.29   &          79.01$\pm$0.49            \\ \hline
MO-GAAL   \cite{liu2019generative}                                                                     & 85.69$\pm$0.18 &                     74.07$\pm$0.67                     \\ \hline
\begin{tabular}[c]{@{}c@{}}$(\bullet)$ Original\\GeoTransform72 \cite{golan2018deep}\end{tabular}                                                                            & 98.60$\pm$0.23                                          & 85.63$\pm$1.48                 \\ \hline
GeoTransform72                     & 98.78$\pm$0.26                  & 87.33$\pm$1.24           \\ \hline
GeoTransform99                      & 99.12$\pm$0.04          &    90.80$\pm$0.61            \\ \hline
\begin{tabular}[c]{@{}c@{}}$(\square)$ (HiTS) GeoTransform35 \\ $(\blacksquare)$ (ZTF)  GeoTransform29\end{tabular}                                                                            & \textbf{99.20$\pm$0.06}                            & \textbf{91.39$\pm$0.76}                                     \\ \hline\hline
\begin{tabular}[c]{@{}c@{}}Welch's t-test p-value\\ $(\bullet)$ v/s $(\square)$ \& $(\blacksquare)$\end{tabular} &    2.0$\times$10$^{-5}$                                                   &  5.3$\times$10$^{-7   }$                                  \\ \hline
\end{tabular}
}
\end{table}

\bibliographystyle{IEEEtran}
\bibliography{IEEEabrv, references}
\end{document}